\documentclass{article}
\usepackage{spconf,amsmath,graphicx}
\usepackage{algorithm}
\usepackage{algpseudocode}
\usepackage[table, scgnames]{xcolor}
\usepackage{tabulary,multirow,overpic,xcolor}
\usepackage{tikz}
\usepackage{tikz,pgfplots}
\usepackage{subfig}
\pgfplotsset{compat = 1.3,
	legend style={font=\scriptsize},
	legend cell align={left},
	legend style={cells={align=left}, draw=black!20},
	grid=both,
	grid style={dotted},
	tick style={draw=none},
	enlarge x limits=false,
	enlarge y limits=false,
	axis line style={draw=black!100},
	axis lines=left,
}

\pgfplotscreateplotcyclelist{mycolormarklist}{%
	chameleon3,mark=diamond*,mark options={solid, fill=chameleon3}\\
	skyblue2,mark=pentagon*,mark options={solid}\\
	scarletred3,mark=triangle*,mark options={solid}\\
	plum1,mark=square*,mark options={solid}\\
	aluminium5,mark=otimes*,mark options={solid}\\
	orange1,mark=square*,mark options={solid}\\
	skyblue3\\
	chocolate2\\
	maroon\\
	butter3\\
}
\pgfplotscreateplotcyclelist{mystylelist}{%
	densely dashdotted\\
	solid\\
	dotted\\
	densely dashed\\
	dashed\\
 dashdotted\\
	densely dotted\\
	densely dash dot dot\\
	dotted\\
	dash dot dot\\
}

\usepackage{soul}
\definecolor{defaultcolor}{gray}{.92}
\definecolor{defaultcolor2}{gray}{.90}
\sethlcolor{defaultcolor2}
\usepackage{amsfonts}
\newcommand{\app}{\raise.17ex\hbox{$\scriptstyle\sim$}}
\def\x{$\times$}

\definecolor{carmine}{rgb}{0.59, 0.0, 0.09}

\definecolor{demphcolor}{RGB}{144,144,144}

\definecolor{xycolor}{RGB}{60, 120, 216}
\definecolor{xycolor}{HTML}{0071bc}

\definecolor{wcolor}{RGB}{103, 78, 167}

\definecolor{dcolor}{RGB}{166, 77,21}

\definecolor{gcolor}{RGB}{204, 102, 153}

\definecolor{tcolor}{RGB}{80, 200, 180}

\definecolor{eicolor}{RGB}{153, 51, 102}

\newcolumntype{x}[1]{>{\centering\arraybackslash}p{#1pt}}
\newcolumntype{y}[1]{>{\raggedright\arraybackslash}p{#1pt}}
\newcolumntype{z}[1]{>{\raggedleft\arraybackslash}p{#1pt}}
\newlength\savewidth\newcommand\shline{\noalign{\global\savewidth\arrayrulewidth
		\global\arrayrulewidth 1pt}\hline\noalign{\global\arrayrulewidth\savewidth}}
\newcommand{\tablestyle}[2]{\setlength{\tabcolsep}{#1}\renewcommand{\arraystretch}{#2}\centering\footnotesize}

\usepackage[pagebackref,breaklinks,colorlinks]{hyperref}

\usepackage[capitalize]{cleveref}
\crefname{section}{Sec.}{Secs.}
\Crefname{section}{Section}{Sections}
\Crefname{table}{Table}{Tables}
\crefname{table}{Tab.}{Tabs.}

\newcommand{\real}{\mathbb{R}}

\newcommand{\bh}{\mathbf{h}}

\newcommand{\bW}{\mathbf{W}}

\newcommand{\bU}{{\mathbf{U}}}

\newcommand{\bV}{{\mathbf{V}}}

\newcommand{\bQ}{{\mathbf{Q}}}

\def\x{{\mathbf x}}

\title{MeMSVD: Long-range temporal structure capturing \\ using incremental SVD}
\name{Ioanna Ntinou$^{1}$ \qquad Enrique Sanchez$^{2}$ \qquad Georgios Tzimiropoulos$^{1,2}$}
\address{$^1$Queen Mary University London, UK  \qquad $^2$Samsung AI Center Cambridge, UK}

\begin{document}
\maketitle

\begin{abstract}
This paper is on long-term video understanding where the goal is to recognise human actions over long temporal windows (up to minutes long). In prior work, long temporal context is captured by constructing a long-term memory bank consisting of past and future video features which are then integrated into standard (short-term) video recognition backbones through the use of attention mechanisms. Two well-known problems related to this approach are the quadratic complexity of the attention operation and the fact that the whole feature bank must be stored in memory for inference. To address both issues, we propose an alternative to attention-based schemes which is based on a low-rank approximation of the memory obtained using Singular Value Decomposition. Our scheme has two advantages: (a) it reduces complexity by more than an order of magnitude, and (b) it is amenable to an efficient implementation for the calculation of the memory bases in an incremental fashion which does not require the storage of the whole feature bank in memory. The proposed scheme matches or surpasses the accuracy achieved by attention-based mechanisms while being memory-efficient. Through extensive experiments, we demonstrate that our framework generalises to different architectures and tasks, outperforming the state-of-the-art in three datasets. 
\end{abstract}

%%%%%%%%% BODY TEXT
\section{Introduction}
\label{sec:intro}

\begin{figure}[t!]
\begin{center}
   \includegraphics[width=0.70\linewidth]{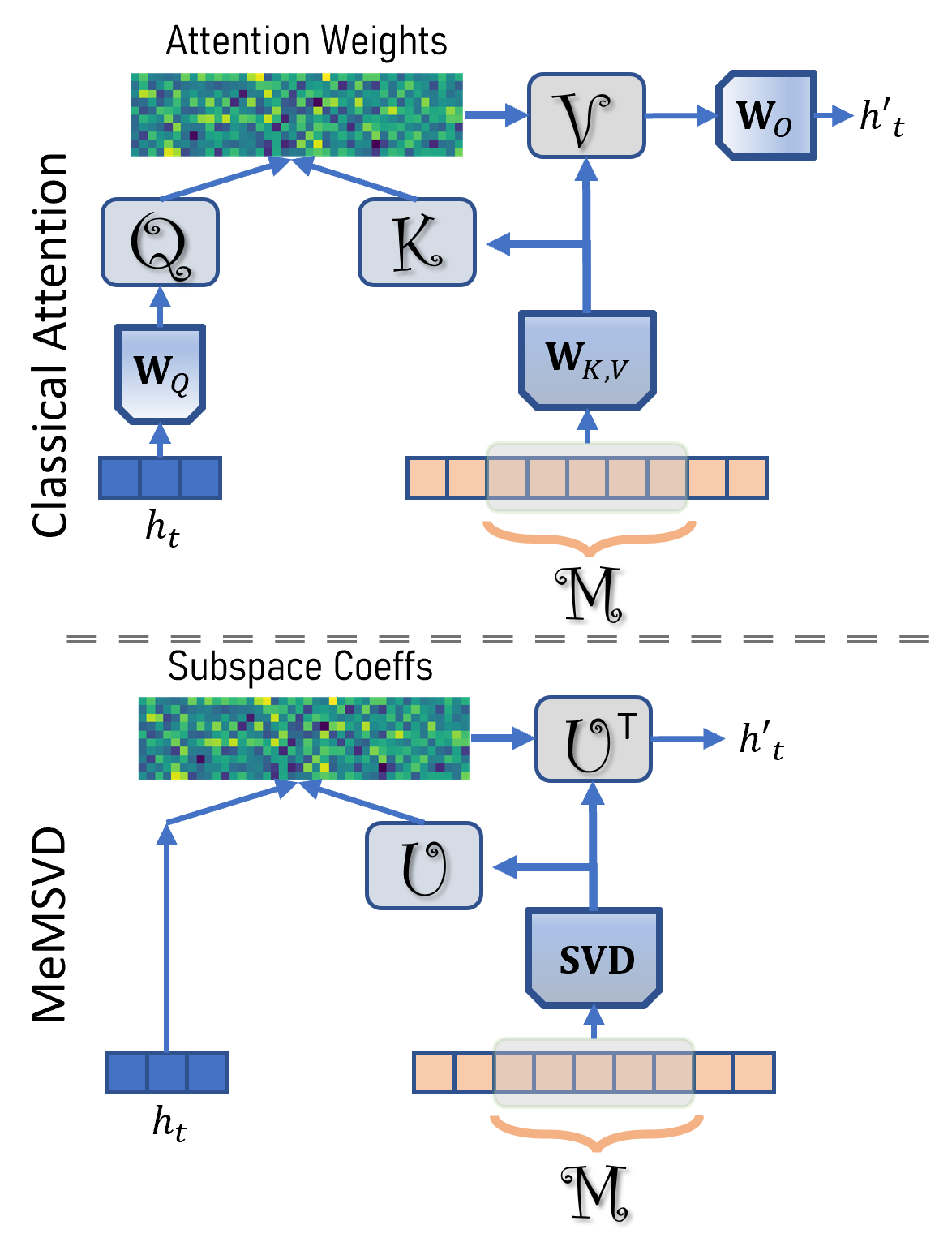}
\end{center}
   \caption{{\bf Top:} Attention-based approaches for memory banks project the clip features $\mathbf{h}_{t}$ into a memory bank through learnable projections. {\bf Bottom:} We propose to compress the memory bank using SVD by projecting and reconstructing the features $\mathbf{h}_{t}$ into the subspace spanned by the memory basis. \vspace{-3pt}}
\label{fig:graphabs}
\end{figure}

Video Action Recognition is a challenging task that requires a proper understanding of spatiotemporal context. Recent advances primarily address trimmed videos with single actions spanning over a few seconds~\cite{wu2021}. However, there's a computational limit on the number of frames a video model can process, thus limiting the temporal receptive field. While this is not a problem for trimmed scenarios, it is an open question how to efficiently incorporate the temporal context in the untrimmed scenarios where long-range understanding is necessary, such as that of e.g. AVA~\cite{gu2018}. This paper focuses on this particular problem, i.e. the effective integration of long-term temporal context in video recognition.

Following the seminal work of~\cite{wu2019}, memory banks have become the standard approach for integrating temporal context into recognition. The motivation behind a memory bank is to provide a model with the capacity to attend to longer temporal support than that the backbone can process at a single forward pass. A memory bank represents the ``supportive information extracted over the entire span of a video''~\cite{wu2019} and is constructed by extracting video features from past and future video clips. Using standard attention mechanisms, the video features from the current clip can be updated according to the memory bank, dramatically increasing performance~\cite{wu2019,pan2021}. However, a well-known limitation of attention-based methods is the increased complexity w.r.t. the number of attended clips, which in turn influences the computational complexity.

In this work, we aim to improve both the memory and the computational efficiency of the long-term feature bank using a new memory representation based on a low-rank approximation of the feature bank via Singular Value Decomposition (SVD). Our motivation stems from the way that deep learning video recognition approaches perform classification: non-linear feature extraction using a video backbone followed by linear classification. Given this, and since the video feature from the current clip interacts with the memory just before the classifier, the following research question arises: \textit{is there any need at all to use non-linear cross-attention between the video feature and the memory bank?}

\noindent\textbf{Key contributions:} We demonstrate that non-linear attention can be entirely omitted for feature-memory interaction without sacrificing accuracy. This, in turn, has 3 important implications: (1) We can derive a condensed memory representation using SVD; (2) We show that projection and reconstruction using the SVD bases is a computationally efficient operation analogous to attention (more than an order of magnitude more efficient), and (3) we observe that a memory representation based on SVD can be \textit{updated} rather than fully recomputed using incremental SVD~\cite{brand2006}, which further optimises the computational complexity of our proposed SVD-based memory representation. Based on (1)-(3), we propose MeMSVD (\cref{fig:graphabs}), a new memory and computationally efficient memory representation for long-term video recognition seamlessly amenable to incremental updates. Through extensive experiments, we demonstrate that (a) our framework generalises to different video architectures and tasks, and importantly, (b) matches or even surpasses the accuracy achieved by the attention-based mechanisms.

%%%%%%%%% Related Work  
\section{Related Work}

\paragraph*{Spatio-temporal Action Detection}  aims at both localising and recognising atomic actions in video clips. Generally, existing research can be divided into methods that seek for efficient architectures that can perform both tasks at low complexity~\cite{Ntinou_2024_CVPR}, and those that focus on the classification task~\cite{wu2019,tang2020, fan2021, li2022}. Despite the recent advances of those exploring the former, their performances are still sub-par w.r.t. those that exploit efficient architectures for classification only. 

\paragraph*{Long term video modelling} has significantly benefited video recognition tasks.  How to tackle such modeling often depends on the task at hand: for trimmed videos there are works that propose consensus blocks across sparsely sampled frames~\cite{lin2019tsm}, or works that propose hierarchical sampling~\cite{ zhou2018}; for untrimmed videos such as those found in AVA the trend is to build on using a compressed memory to augment the temporal support~\cite{wu2019,wu2022,pan2021,tang2020}. Wu et al. \cite{wu2019} proposed to use long-term feature banks (LFB) to provide temporal supportive information up to 60s for computing long-range interaction between actors. Feature banks allow saving actor features at a given temporal position to further use it later in the scene. Most recent and sophisticated methods like \cite{pan2021, tang2020}, combine memory banks with spatio-temporal attention mechanisms to leverage interaction spatially and temporally. MeMViT~\cite{wu2022} builds on memory-like design that allows for single-stage training.

\paragraph*{Softmax free transformers} refers to the line of research that investigates ways to drop the softmax operation which is an expensive operation from  attention-based algorithms. SOFT \cite{lu2021} proposes the use of a Gaussian kernel function. Hamburger~\cite{geng2021} explores the use of learnable low-rank Matrix Decompositions as an approximation to self-attention. In this paper, we propose the use of SVD as a parameter-free attention mechanism to a memory bank.

\section{Method} \label{sec:method}
\subsection{Preliminaries} \label{ssec:prel}

\paragraph*{Problem definition:} Spatio-temporal action recognition in video sequences aims at localising and recognising actions performed by all actors in a given video clip. A long video (e.g. several minutes long) is split into short consecutive clips each spanning a fixed number of frames (e.g. 64). At time $t$ is processed by a video backbone (e.g. SlowFast~\cite{feichtenhofer2019}). To compute actor-specific features, a region-based person detector (e.g., Faster R-CNN~\cite{ren2015}) is applied to the central clip frame, generating $N_t$ region proposals. These proposals are used to align the backbone features through RoI pooling, and then are flattened to generate an actor feature $\textbf{h}_{t,s} \in \mathbb{R}^{1 \times d},\;s=1,\dots,N_t$. The actor feature is fed to a classification head $\Phi$ to compute class probabilities for the clip or actor. 

\paragraph*{Memory construction:} In short-term action recognition, each actor feature $\textbf{h}_{t,s}$ is classified independently. In long-term action recognition, a memory feature bank is constructed to encode information from all clips (and all actors) in a temporal window. The main goal of adding a feature bank is for the current clip features to \textit{look} beyond its temporal support. A standard approach consists of first populating the memory bank with a pre-trained backbone. Specifically, we define $\textbf{H}_{t} \in \real^{N_t \times d}$ as the concatenation of all $\textbf{h}_{t,s}$ for $s=1,\dots,N_t$. Then, a memory feature bank over a temporal window $[t-w, t+w]$ is defined as the concatenation of clip actor features: $\mathbf{M}=[\textbf{H}_{-w},\dots,\textbf{H}_{w}] \in \real^{N_{mem} \times d}$, with $N_{mem} = \sum_{t'=t-w}^{t+w}N_t$.

\paragraph*{Memory interaction with attention:} A cross-attention layer is used to model the interaction between the actor feature $\mathbf{h}_{t,s}$ and the memory bank $\mathbf{M}$. The keys and value matrices $\mathbf{K}_{mem}, \mathbf{V}_{mem} \in \real^{N_{mem} \times d_u}$ are formed from $\mathbf{M}$ using projection layers $\bW_k \in \real^{d \times d_u}$, $\bW_v \in \real^{d \times d_u}$, whereas the query token $\mathbf{q}_{t,s}\in \real^{ 1 \times d_u}$ is computed from the actor feature $\mathbf{h}_{t,s}$ using a projection layer $\bW_q$. Denoting by $\mathbf{K}_{mem}, \mathbf{V}_{mem}$ as key and value matrices, and by $\mathbf{q}_{t} \in \real^{ 1 \times d_u}$ the query, the updated attended features are computed as:
\begin{equation}
\begin{aligned}
\mathbf{h}' &= {\rm{Softmax}}(\mathbf{q}_{t,s} \cdot \mathbf{K}_{mem}^T / \sqrt{d}) \mathbf{V}_{mem} , \\
\bh_{t,s} &= \bh_{t,s} + \mathbf{h}' \bW_o 
\label{eq:xattn}
\end{aligned}
\end{equation}
where $\bW_o \in \real^{d_u \times d}$ is used to project the output of the attended features back to the original dimension. 

\subsection{Memory Attention as a Subspace Projection}
\label{ssec:svd}
We propose a computationally efficient alternative to the memory plus attention set described in \cref{ssec:prel} as follows: 
\paragraph*{Subspace Memory:} Firstly, we propose to obtain a condensed memory representation by computing the subspace spanned by the rows of $\mathbf{M}$ (or the columns of $\mathbf{M}^T$) using SVD according to which $\mathbf{M}^T$ can be decomposed as:
\begin{equation}
    \mathbf{M}^T = \mathbf{U}\mathbf{\Sigma}\mathbf{V}^T.  
\end{equation}
with $\bU \in \mathbb{R}^{d \times N_{mem}}$\footnote{We assume $N_{mem} \le d$}, $\mathbf{\Sigma} \in \mathbb{R}^{N_{mem} \times N_{mem}}$ and $\bV \in \mathbb{R}^{N_{mem} \times N_{mem}}$. By keeping the top $n_c$ components of $\mathbf{U}$, a set of the most significant basis vectors can be obtained as $\mathbf{U}_{mem}=\mathbf{U}^T[:n_c,:] \in \real^{n_c\times d}$.  Note that $n_c \ll N_{mem}$ (typically $20-30x$ smaller) offering memory savings.

\paragraph*{Subspace projection:} Secondly, we propose projection and reconstruction onto $\mathbf{U}_{mem}$ as an alternative to the cross-attention scheme of Eq.~\ref{eq:xattn}:
\begin{equation}
%\begin{aligned}
\mathbf{h}' = \left[\mathbf{h}_{t,s} \cdot \mathbf{U}_{mem}^T \right] \mathbf{U}_{mem} , \Longrightarrow
\bh_{t,s} = \bh_{t,s} + \mathbf{h}'. 
\label{eq:proj}
%\end{aligned}
\end{equation}
As known from linear algebra, bases can be expressed as a linear combination of the data i.e. $\mathbf{U}_{mem}=\mathbf{C}\mathbf{M}$, where $\mathbf{C}\in \real^{n_c\times N_{mem}}$ is some coefficient matrix. Hence:
\begin{equation}
\begin{aligned}
\mathbf{h}' &= \left[\mathbf{h}_{t,s} \cdot \mathbf{U}_{mem}^T \right] \mathbf{U}_{mem} , \\
  &= \mathbf{h}_{t,s} \cdot \mathbf{M}^T \mathbf{C}^T \mathbf{C}\mathbf{M} 
   = \mathbf{a}_{t,s} \cdot \mathbf{C}^T \mathbf{C}\mathbf{M}.
\label{eq:proj2}
\end{aligned}
\end{equation}
The above equation clearly shows $\mathbf{h}_{t,s}$ firstly interacts with the memory features in $\mathbf{M}$ to give interaction coefficients 
$\mathbf{a}_{t,s}$ (akin to attention) and finally produces as output a linear combination of the memory features (also akin to attention). Hence, we can interpret the projection and reconstruction as a mechanism analogous to memory attention. 

Two notable differences of the above formulation with the attention-based formulation of Eq.~\ref{eq:xattn} are: (a) there is no non-linearity (e.g. softmax non-linearity), and (b) the computational cost is reduced from $\mathcal{O}(N_{mem}d)$ to $\mathcal{O}(n_cd)$ ($n_c \ll N_{mem}$). In the general scenario where SVD is computed through the eigendecomposition of the covariance matrix, the computational cost of the SVD is $\mathcal{O}(N_{mem}^2d)$ which is similar to that of the projection operations in \cref{eq:xattn}. However, newer methods for SVD approximation based on randomised QR decompositions~\cite{halko2009} can be used when the target rank (i.e. the number of components $n_c \ll N_{mem}$) is known. In such case, the cost of computing the SVD basis is reduced to $\mathcal{O} (n_c N_{mem} d)$ (~\cite{halko2009}, Algorithm 5.1).  Finally, note that our method reduces the complexity in a factor of $n_c/N_{mem}$, and, more importantly, our proposed method \textit{does not require the learning of projection matrices to produce the attention weights}. This in turn offers a significant advantage in terms of FLOPs and number of parameters that as we shall see comes with no drop in accuracy.

\subsection{Online SVD for long-term video understanding}
\label{ssec:svd}
In long-term video understanding, it is expected that queries to the memory bank will be sequential. In such a scenario, computing the SVD basis at each time step can be redundant as the temporal window for the memory bank shifts by a single time step. The memory at time $t$ differs from that at time $t-1$ in one clip only. In this setting, we can further reduce the complexity of our method by resorting to updating the SVD basis in a rolling manner, i.e. with one clip at a time. In other words, rather than keeping a full representation of the memory bank at time $t$ $\mathbf{M}_t=[\textbf{H}_{-w},\dots,\textbf{H}_{w}] \in \real^{N_{mem} \times d}$, we can keep only the basis $\{\mathbf{U}_{mem}, \mathbf{\Sigma}_{mem}\} \in \mathbb{R}^{n_c \times (d+1)}$ and update it with the newest clip $\textbf{H}_{w+1} \in \mathbb{R}^{N_{w+1} \times d}$, composed of the $N_{w+1}$ actors in clip $w+1$. How to efficiently update an SVD model is a well-studied problem~\cite{brand2006}, which boils down to rotating the bases to align them to the orthogonal elements of the updating vectors. Using \cref{alg:cap} (see \cite{brand2006} for a formal derivation), we can update the basis $\bU_{mem}' \in \mathbb{R}^{n_c \times d}$ from a new set of actor features $\mathbf{H}_{w+1} \in \mathbb{R}^{N_{w+1} \times d}$, without the need of computing again the full SVD. A ``forgetting factor'' $\lambda \in (0,1)$ is included to control how the existing basis will be influenced by the future clips.

\begin{algorithm}
\caption{Updating the SVD basis}
\label{alg:cap}
\textbf{Input}: $\textbf{H}_{w+1} \in \mathbb{R}^{N_{w+1} \times d}$,$\{\mathbf{U}_{mem}, \mathbf{\Sigma}_{mem}\} \in \mathbb{R}^{n_c \times (d+1)}$ \\
$\lambda$ forgetting factor
\begin{algorithmic}[1]
\State  $\hat{\mathbf{H}}_{w+1} = \mathbf{H}_{w+1} ({\bf I} - \bU_{mem}^T \bU_{mem})$

\State $\textbf{Q}\textbf{R} \xleftarrow[]{QR} \hat{\mathbf{H}}_{w+1}^T $ 

\State $\mathbf{U}' \mathbf{\Sigma}' \mathbf{V}'^T \xleftarrow[]{SVD} \begin{bmatrix} \lambda \mathbf{\Sigma}_{mem} &\textbf{U}_{mem} \mathbf{H}_{w+1}^T \\
\textbf{0} &\textbf{R}
\end{bmatrix}$

\State $\begin{matrix} \bU_{mem}^T \leftarrow \begin{bmatrix}  \bU_{mem}^T & \bQ\end{bmatrix} \bU'; & \mathbf{\Sigma}_{mem}' \leftarrow \mathbf{\Sigma}';\end{matrix}$

\end{algorithmic}
\textbf{Output}: updated $\{\mathbf{U}_{mem}, \mathbf{\Sigma}_{mem}\} \in \mathbb{R}^{n_c \times (d+1)}$
\end{algorithm}

Note that the SVD in Step 3 in \cref{alg:cap} is computed over the upper-triangle matrix of $(n_c + N_{w+1})\times (n_c + N_{w+1})$ components, thus incurring in negligible complexity. In the extreme case where $\mathbf{H}_{w+1}$ is composed of a single feature vector only (e.g. when the clip contains one actor only), the QR decomposition reduces to just $\mathbf{R} = r = \| \mathbf{H}_{w+1} \|_2$, and $\mathbf{Q} = \mathbf{H}_{w+1} r^{-1}$. Overall, the computational complexity of updating the SVD basis reduces to $\mathcal{O}(n_c^2)$. Additionally, keeping an online SVD basis results in lower memory consumption as it eliminates the need to store all clips in memory. While adding video clips to the memory basis in an online fashion is an appealing property, we note that it can lead to long-term divergence between the actual eigenvectors that would be computed from the full memory at time $t$ and those updated throughout a video processing. We analyse the effect of computing the incremental SVD in \cref{experiments}.

\section{Experiments}
\label{experiments}
We conduct extensive experiments with three architectures building both on CNN-based feature representations~\cite{feichtenhofer2019, pan2021,fan2020} and vision transformers~\cite{fan2021,li2022,wu2022}. In particular, we experiment with a) SlowFast~\cite{feichtenhofer2019} b) ACAR-Net~\cite{pan2021} and c) MViTv2-S~\cite{fan2021,li2022,wu2022}. For each architecture, we create a \textbf{memory bank} by storing for each actor in a clip actor-specific features. For ablation studies we use the AVA v2.2 ~\cite{gu2018}. Our goal in this paper is not to develop a novel architecture that attains state-of-the-art results but to show that our proposed attention to a memory bank is more efficient than the equivalent architectures of those methods.

\subsection{Experimental Setup}
\label{ssec:implementation}

\paragraph*{AVA dataset.} AVA is a spatio-temporal action localisation dataset containing $299$, 15-minute long videos, segmented into 1-second clips. Each clip is labelled with a 2D bounding boxing for each person appearing in the clip along with a multi-label annotation specifying which actions the person of reference is engaged in. We follow the standard evaluation protocol, reporting the mean Average Precision (mAP) on 60 classes \cite{gu2018}. For the detection task, we follow ~\cite{wu2022, fan2021} and use the pre-computed bounding boxes of \cite{wu2019}.

\paragraph*{Training and Inference.} All models are trained following the corresponding publicly available code with the same learning rate and scheduler as reported by the corresponding authors~\cite{pan2021, li2022, feichtenhofer2019}.  We train the backbone on a clip-level basis to produce offline feature representations and build the memory banks. We then train the model incorporating the attention-to-memory module, using \textit{either} our proposed approach \textit{or} the standard non-local block, i.e. we remove the Non-Local block on SlowFast and ACAR-Net when training the model with our proposed method. For inference, we sample a clip of $T$ frames using a stride $\tau$ centred at the frame that is to be evaluated. For the offline version of the MeMSVD (i.e. without the incremental steps), we fetch features that correspond to a temporal window of $t$ seconds and pass them through our model as well. For the online version of the MeMSVD, oMeMSVD,  we need to process clips \textit{sequentially}. 

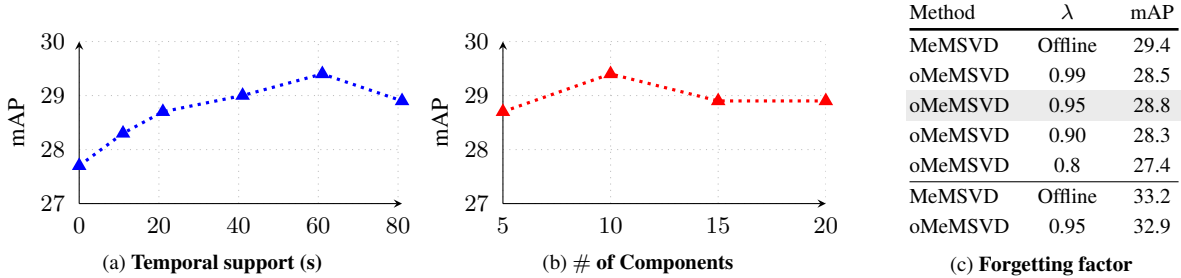
\begin{figure*}[t!]
\small
   \begin{minipage}{.6\textwidth}
  \centering

			\subfloat[\textbf{Temporal support (s)}\label{tab:abl:temp}]{%
				\begin{tikzpicture}
				\begin{axis}[
				clip=false,
				width=0.55\textwidth,
				height=0.35\textwidth,
				xlabel={},
				ylabel={mAP},
				xmin=0,
                ymin=27,
				ymax=30,
				every axis plot/.append style={very thick,mark options={scale=0.5, solid}},
				legend style={at={(1.0,1.0)},anchor=north east, cells={align=left}}
				]
				\addplot+[blue,mark=triangle*,mark options={solid}, dotted]
				table[row sep=crcr] {
					x y\\
                        0   27.7 \\
					11	28.3\\
					21	28.7\\
					41	29.0\\
					61	29.4\\
					81	28.9\\
				};
				\end{axis}
				\end{tikzpicture}
			}
			\subfloat[\textbf{$\#$ of Components}\label{tab:abl:components}]{%
				\begin{tikzpicture}
				\begin{axis}[
				clip=false,
				width=0.55\textwidth,
				height=0.35\textwidth,
				xlabel={ },
				ylabel={mAP},
				xmin=5,
                ymin=27,
                ymax=30,
				every axis plot/.append style={very thick,mark options={scale=0.5, solid}},
				legend style={at={(1.0,1.0)},anchor=north east, cells={align=left}}
				]
				\addplot+[red,mark=triangle*,mark options={solid}, dotted]
				table[row sep=crcr] {
					x y\\
					5	28.7\\
					10	29.4\\
					15	28.9\\
					20	28.9\\
				};

				\end{axis}
				\end{tikzpicture}
			} 
   \end{minipage} 
     \begin{minipage}{0.35\textwidth}
     \centering
				\subfloat[\textbf{Forgetting factor }\label{tab:abl:forgetting}]{%
		\tablestyle{1.0pt}{1.20}
		\begin{tabular}{@{} 
				lx{40}x{20}
				@{}}
			Method & $\lambda$  & mAP\\
			\shline
           MeMSVD  & Offline   & 29.4 \\
           oMeMSVD & 0.99  & 28.5\\
    		 \rowcolor{defaultcolor}
           oMeMSVD &0.95  & 28.8 \\
           oMeMSVD &0.90   & 28.3 \\
           oMeMSVD & 0.8   & 27.4 \\
            \hline
           MeMSVD & Offline   & 33.2 \\
           oMeMSVD & 0.95  & 32.9\\

		\end{tabular}\vspace{-4pt}
	%}  %
			}
    \end{minipage}

		%}%\vspace{-2mm}
		\caption{\textbf{Ablation Experiments.} We conduct ablation on (a): temporal support of the memory features, (b): number of components of the memory representation, and (c): forgetting factor. As backbone we have an ACARNet-50 $8 \times 8$  (upper cell) and  ACARNet-101 $8 \times 8$ (bottom cell). All results are on conducted on the AVAv2.2 dataset~\cite{gu2018} with Kinetics-400~\cite{kay2017} or Kinetics-700\cite{smaira2020} pre-training. The \hl{gray row} denotes default choice. (mAP in \%)}\label{fig:tradeoff}
	\end{figure*}

\subsection{Ablation Experiments}

For our ablation studies, we use  ACAR-Net~\cite{pan2021} architecture with the SlowFast-50 $8 \times 8$. We follow the same training procedure as that of \cite{pan2021} to train our models, by replacing their Non-Local block by our MeMSVD approach. Using this setting, we study a) the impact of the temporal support, b) the impact of the memory compression by means of kept components, and c) the impact of the forgetting factor in the performance of the online SVD w.r.t. the offline version.

\paragraph*{Temporal Support.} 
We first analyse the effect of the temporal support in our memory features. As a baseline, we re-train the ACAR-Net without memory, and use the subsequent features to train a variety of models where the SVD bases are computed over a memory bank spanning a different time length. In \cref{tab:abl:temp} we observe that a significant gain is achieved when considering a spanning window of $61$ seconds, centred at the target clip. We also note that \textit{increasing the temporal support incurs an insignificant additional computational cost}. For the rest of our ablation studies, we fix the window size to $61$ seconds due to the strong performance.

\paragraph*{Number of Components.}
We investigate the impact of using memory representations of varying rank. For both training and inference  we experiment with the number of components $n_{c}$ and, hence, the rank of the memory base $\bU_{mem}$. From \cref{tab:abl:components} we observe that keeping a vector base of size $10 \times  d $ is enough to reach mAP $29.4$ on AVA v2.2. 

\paragraph*{Forgetting Factor.}
We now evaluate the impact of applying the online SVD on the performance of our method, as well as the effect of the forgetting factor. For both training and evaluation, we perform sequential reading of the clips to maintain the temporal coherence of the memory bases. We incrementally update the memory by fetching only the features that are requested for the next clip (i.e. at time $t+30$), and perform the memory update as described in \cref{ssec:svd}. Incremental updates happen every second when a new clip arrives. As an ablation, we experiment with different values of the forgetting factor. A forgetting factor of $\lambda = 0$ denotes that the new memory representation is made only from new observations while a forgetting factor of $\lambda = 1$ assumes that we maintain all observations as time progresses. 

We use the ACAR-Net architecture with SlowFast-50 and SlowFast-101 backbones. We refer to the online version of our method by oMeMSVD, and we make comparisons w.r.t. its offline counterpart. We train and evaluate our pipeline using a forgetting factor that ranges from $0.8$ to $0.99$. The results on AVA are shown in \cref{tab:abl:forgetting}. We observe that heavily down-weighting the effect of early observations negatively impacts the performance (e.g. when $\lambda=0.8$), validating the necessity for long-temporal support. A factor of $\lambda = 0.95$ achieves a performance that is close to that of the offline version.

\paragraph*{Generalisation to different backbones.}  We assess our approach's performance using various backbones, including ACAR-Net~\cite{pan2021} (ResNet-50 and ResNet-101), MViTv2-S $16 \times 4$~\cite{li2022}, and SlowFast~\cite{feichtenhofer2019} (ResNet-50 and ResNet-101). Initially, we retrain these backbones to generate feature representations and construct a memory bank. Then we retrain the models, comparing those with attention to memory and our proposed MeMSVD models, ensuring a fair evaluation using the same pre-computed feature bank.

As shown in \cref{table:backbones}, MeMSVD produces consistent accuracy gains of over $2\%$ on top of state-of-the-art video modelling methods, including ACAR-Net with SlowFast R50 $8 \times 8$, and SlowFast R101 $8 \times 8$. Note that we achieve these improvements with a significant parameter reduction compared to attention-based methods. It is important to remark that our method has even fewer parameters and FLOPs than the backbone for CNN-based methods, as it does not include the Non-Local blocks which attend to the spatial features to provide the actor-actor and actor-to-memory interactions. 

\begin{table*}[t!] 
\small
\setlength{\tabcolsep}{8pt}
\caption{Comparison of accuracy vs complexity tradeoff for different architectures using both our proposed MeMSVD and the default attention-to-memory approach. Our method has notable gains with a significant reduction in complexity. \vspace{-7pt}}
\label{table:backbones}
\begin{center}
\begin{tabular}{c|c|c|c|c|c} 
\multicolumn{1}{c|}{\bf Architecture} & \multicolumn{1}{c|}{\bf Method}  &\multicolumn{1}{c|}{\bf Pretrain} & \multicolumn{1}{c|}{\bf mAP} &\multicolumn{1}{c|}{\bf FLOPs (G)} &\multicolumn{1}{c}{\bf Param (M)} \\
 \hline
\multirow{3}{*}{SlowFast-50 $8 \times 8$~\cite{feichtenhofer2019}} & Backbone &  \multirow{3}{*}{ K400}  & 26.2 & 97.9& 54.0  \\
& Attention &  &  26.8 & 112.0  & 54.0 \\
\rowcolor{defaultcolor} 
& {\bf MeMSVD}  &  & {\bf 27.0} & {\bf 97.6}  & {\bf 35.1}\\
 \hline
 \multirow{3}{*}{SlowFast-101 $8 \times 8$~\cite{feichtenhofer2019}} & Backbone  &  \multirow{3}{*}{ K700} & 29.2 & 158.2& 78.1\\
& Attention  & & 30.5 & 172.3& 78.1\\
\rowcolor{defaultcolor} 
& {\bf MeMSVD} & & {\bf 30.9} & {\bf157.9} & {\bf 63.4}\\
\hline
\multirow{3}{*}{ACAR-Net $8 \times 8$ R50~\cite{pan2021}} & Backbone   &  \multirow{3}{*}{ K400} & 27.7& 98.6& 56.9\\
& Attention &  & 29.0&  112.6& 56.9  \\
\rowcolor{defaultcolor} 
& {\bf MeMSVD}  &  & {\bf 29.4 }& {\bf 98.2} & {\bf 38.0} \\
\hline

\multirow{3}{*}{ACAR-Net $8 \times 8$ R101~\cite{pan2021}} & Backbone  & \multirow{3}{*}{ K700} & 31.3 &{\bf 158.2} & 81.0 \\
& Attention  & & 32.9 & 172.9  &  81.0\\
\rowcolor{defaultcolor} 
& {\bf MeMSVD}  & & {\bf 33.2} & 158.5 & {\bf 66.3} \\
\hline
\multirow{3}{*}{MViTv2-S $16\times4$~\cite{li2022}} & Backbone & \multirow{3}{*}{ K400}  & 28.2 &  {\bf 64.4} & {\bf 34.4} \\
& Attention &                         & {\bf 30.1} &  77.4  &  55.5\\
\rowcolor{defaultcolor} 
& {\bf MeMSVD} &                         & 30.0 &  {\bf 64.4}  & {\bf 34.3} \\
\hline
\end{tabular}
\end{center}
\end{table*}

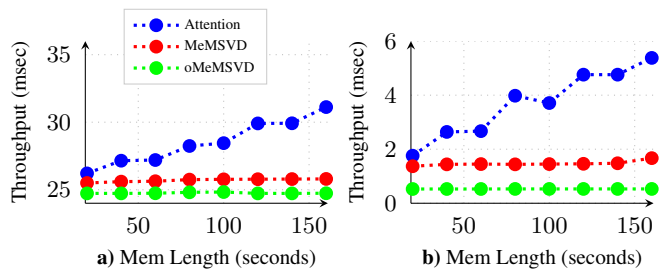
\begin{figure}[b!] %\vspace{-1mm}
		%\resizebox{1\textwidth}{!}{
			%
   \small
   \begin{minipage}{.6\textwidth}
  %\centering

			\subfloat{%
				\begin{tikzpicture}
				\begin{axis}[
				clip=false,
				width=0.45\textwidth,
				height=0.35\textwidth,
				xlabel={ \footnotesize {\bf a)} Mem Length (seconds)},
				ylabel={\footnotesize Throughput (msec)},
				xmin=19,
                xmax=161,
                ymin=24,
				ymax=36,
				every axis plot/.append style={very thick,mark options={scale=0.5, solid}},
				legend style={nodes={scale=0.8, transform shape}, at={(0.75, 1.2)},anchor=north east, cells={align=left}}
				]
				\addplot+[blue,mark=*,mark options={solid}, dotted]
				table[row sep=crcr] {
					x y\\
                        20  26.21 \\
					40	27.15\\
					60	27.20\\
					80	28.24\\
					100	28.45\\
                        120 29.91 \\
                        140 29.92 \\
                        160 31.12 \\
				};
    \addplot+[red,mark=*,mark options={solid}, dotted]
				table[row sep=crcr] {
					x y\\
                       20  25.50 \\
					40	25.60\\
					60	25.63\\
					80	25.75\\
					100	25.77\\
                       120 25.78 \\
				140	25.79\\
				160	25.80\\
					%180	25.81\\
					%200	25.86\\
				};
    \addplot+[green,mark=*,mark options={solid}, dotted]
				table[row sep=crcr] {
					x y\\
                        20  24.73 \\
					40	24.73\\
					60	24.74\\
					80	24.81\\
					100	24.83\\
                        120  24.73 \\
					140	24.73\\
					160	24.74\\
					%180	24.81\\
					%200	24.83\\
				};
    \legend{Attention, MeMSVD, oMeMSVD}
    %\legend{Attention, Ours}
				\end{axis}
				\end{tikzpicture}
			}
			\subfloat{%
				\begin{tikzpicture}
				\begin{axis}[
				clip=false,
				width=0.45\textwidth,
				height=0.35\textwidth,
				xlabel={ \footnotesize {\bf b)} Mem Length (seconds)},
				ylabel={\footnotesize Throughput (msec)},
				xmin=19,
                xmax=161,
                ymin=0,
				ymax=6,
				every axis plot/.append style={very thick,mark options={scale=0.5, solid}},
				legend style={at={(0.35,1.0)},anchor=north east, cells={align=left}}
				]
				\addplot+[blue,mark=*,mark options={solid}, dotted]
				table[row sep=crcr] {
					x y\\
                        20  1.76 \\
					40	2.64\\
					60	2.67\\
					80	3.98\\
					100	 3.71\\
                        120 4.76 \\
                        140  4.76 \\
                        160  5.38\\
				};
        \addplot+[red,mark=*,mark options={solid}, dotted]
				table[row sep=crcr] {
					x   y\\
                        20   1.37\\
					40	 1.44\\
					60	 1.45\\
					80	 1.44\\
					100	 1.45\\
                        120  1.46\\
				    140	 1.48\\
				    160	 1.67\\
				};
    \addplot+[green,mark=*,mark options={solid}, dotted]
				table[row sep=crcr] {
					x y\\
                        20 0.527  \\
					40	0.528\\
					60	0.528\\
					80	0.53\\
					100	0.53\\
                        120   0.53\\
					140	0.53\\
					160	0.53\\
				};
   % \legend{Attention,  MeMSVD, oMeMSVD}
				\end{axis}
				\end{tikzpicture}
			} 
   \label{Table:b}
   \end{minipage} 
     
		%}%\vspace{-2mm}
  \vspace{-2mm}
	\caption{a) throughput time for the \textbf{full network} and b) for the \textbf{head } only using an MViTv2-S $16 \times 4$~\cite{li2022} backbone with three different heads; an attention to memory module (blue), MeMSVD (red) and oMeMSVD (green). We assumed a fixed number of 3 actors per clip in the memory. Results are measured on a single NVIDIA 3090 GPU.}\label{fig:throughtput} \vspace{-5mm}
	\end{figure}

\paragraph*{Throughput time} 
 We measure throughput for a fixed MViTv2-S $16 \times 4$~\cite{li2022} backbone, using the attention-to-memory module and our proposed MeMSVD and oMeMSVD methods, for a varying window length spanning $0-160$ msec. We report average throughput time for both the full network and the head only, in Fig.~\ref{fig:throughtput}. We note that both MeMSVD and oMeMSVD get a significant gain in speed w.r.t. attention-based methods as the size of the memory increases. Our proposed MeMSVD with and without incremental updates remain almost stable across different window lengths, hence allowing the use of very large temporal windows.

\begin{table}[h!]
\caption{Comparison with previous work on AVAv2.1~\cite{gu2018}. $^\dagger$ indicates results reproduced by us.} \vspace{-1em}
\label{tab:ava:v21}
\begin{center}
\scalebox{0.80}{%
\begin{tabular}{c|c|c}
\multicolumn{1}{c|}{\bf Method}  &\multicolumn{1}{c|}{\bf Pretrain} &\multicolumn{1}{c}{\bf mAP (\%)} \\
\shline
LFB, R-101+NL~\cite{wu2019} &  \multirow{7}{*}{ K400} & 27.7 \\
ATX, I3D~\cite{girdhar2019} &  & 25.0\\
SlowFast-50 $8 \times 8$~\cite{feichtenhofer2019} &  & 24.8\\
LFB, R-50+NL \cite{wu2019} &  & 25.8 \\
 ACAR-Net SF-50 $8 \times 8$~\cite{pan2021}$^\dagger$ &  & 27.9 \\
 ACAR-Net SF-101 $8 \times 8$~\cite{pan2021}& \multirow{3}{*}{ K700}& 30.0 \\
 \rowcolor{defaultcolor}
 {\bf MemSVD} ACAR SF-50 $8 \times 8$&  & {\bf28.1}  \\
\rowcolor{defaultcolor}
 \textbf{MemSVD}  (MViTv2-S,  $16 \times 4$)  &  & {\bf 29.5}\\
 \hline
ACAR-Net SF-101 $8 \times 8$~\cite{pan2021}$^\dagger$ &  K700&  32.2 \\
 \rowcolor{defaultcolor}
 {\bf MemSVD} (ACAR SF-101 $8 \times 8$) &  & \bf{32.3} \\

\end{tabular}}
\end{center}
\end{table}

\begin{table}[h!]
\small
\caption{Comparison with previous work AVAv2.2~\cite{gu2018}. $^\dagger$ indicates results reproduced by us.} \vspace{-1em}
\label{tab:ava:v22}
\begin{center}
\scalebox{0.80}{%
\begin{tabular}{c|c|c}
\multicolumn{1}{c|}{\bf Method}  &\multicolumn{1}{c|}{\bf Pretrain} &\multicolumn{1}{c}{\bf mAP (\%)}\\
\shline
SlowFast-50 $8 \times 8$~\cite{feichtenhofer2019} & \multirow{7}{*}{K400} & 22.7 \\
WOO, SlowFast-50~\cite{chen2021} &  & 25.4 \\
MViTv2-S,  $16 \times 4$~\cite{li2022}  &  & 27.0\\
MeMViT-16, $16 \times 4$~\cite{wu2022} &  & 29.3 \\
ACAR-Net SF-50 $8 \times 8$ $^\dagger$~\cite{pan2021} &  & 29.0 \\
\rowcolor{defaultcolor}
\textbf{MemSVD} (ACAR SF-50 $8 \times 8$) &  & {\bf 29.4}\\
\rowcolor{defaultcolor}
\textbf{MemSVD}  (MViTv2-S,  $16 \times 4$)  &  & {\bf 30.0}\\
\shline
AIA, SlowFast-101 $8 \times 8$~\cite{tang2020} & \multirow{6}{*}{ K700}  & 32.3 \\
CollabMem,  SF-101~\cite{yang2021}&  & 31.6  \\
MViTv2-B, $32 \times 3$ \cite{li2022} &  &  32.5 \\
MeMViT-B, $32 \times 3$ \cite{wu2022}  &  & {\bf 34.4}\\
ACAR-Net SF-101 $8 \times 8$$^\dagger$~\cite{pan2021} &  & 32.9 \\
\rowcolor{defaultcolor}
\textbf{MemSVD}  (ACAR SF-101 $8 \times 8$) &  &  33.2  \\
\end{tabular}}
\end{center}
\end{table}

\section{Comparison with the state of the art}
\subsection{Action Localisation} 

We analyse our proposed MeMSVD against state-of the art methods for AVA v2.1 (\cref{tab:ava:v21}) and AVA v2.2 (\cref{tab:ava:v22}). We split our results according to the method and the pre-training checkpoints. On AVA v2.1 our proposed approach achieves $28.1$ mAP when using ACAR-Net SF-50 and $29.5$ mAP when using MViTv2-S $16 \times 4$ pre-trained on Kinetics-400. On AVA v2.2 with models pre-trained on Kinetics-400, our MeMSVD gets $29.4$ mAP with ACARNet-50 $8 \times 8$ backbone and $30.0$ with  MViTv2-S $16 \times 4$ backbone, surpassing current state of the art \cite{wu2022}. For models pre-trained on Kinetics-700, MeMSVD with ACARNet-101 surpasses the attention counterpart but it does not reach \cite{wu2022} accuracy.  

\subsection{Action Recognition}
To show the generalisation capabilities of our approach, we extend it to the domain of action recognition in trimmed videos using the Charades dataset~\cite{sigurdsson2016}. It consists of $9.8k$ videos of around $30$ seconds in duration. Following prior work, we report the mean average precision (mAP) over 157 classes on the validation set. To infer the action over the whole video, we follow the standard approach: we sample multiple clips corresponding to different spatio-temporal crops, and apply max-pool over the prediction scores to form the final predictions. To train and test our method we sample clips within a temporal window around the central clip as well as multiple clips from the whole video. Since Charades videos vary in size, we extract $N$ clips from each video, constructing the memory bank with clips of varying stride. To fill the memory bank for our experiments we uniformly sample $30$ temporal crops which are spatially centred. For {\bf MeMSVD} we then keep $5$ components. Following the training schedule proposed by \cite{yang2021}, our framework achieves a mAP of $43.0$, surpassing all prior results with a backbone pre-trained on Kinetics-400, as shown in Table~\ref{tab:sota:charades}.

 \def\x{$\times$}

\begin{table}[h] 
\caption{Comparison with previous work on Charades.} 
\centering
\small
	\tablestyle{2pt}{1.05}
	\begin{tabular}{c|c|c|r|r}
	\multicolumn{1}{c|}{\bf Method} &  \multicolumn{1}{c|}{\bf Pretrain} &  {\bf mAP}   & {\bf\scriptsize FLOPs\x views} & {\bf Param} \\   
	\shline
	LFB +NL \cite{wu2019} &  \multirow{5}{*}{\scriptsize K400}  & 42.5 & 529\x3\x10 & 122    \\
	{SlowFast} {\scriptsize{50, 8\x8}}~\cite{feichtenhofer2019}  &   & 38.0  &  65.7\x3\x10  & 34.0 \\
	{SlowFast} {\scriptsize{101+NL, 16\x8}}~\cite{feichtenhofer2019}  &  & 42.5  &  234\x3\x10  & 59.9  \\
	CollabMemories \cite{yang2021}&  &42.9 & 135 \x 3  \x 10 & N/A\\
	MViT-B, 16\x4 \cite{fan2021}&  & 40.0 & 70.5\x3\x10 & 36.4 \\ 
        \rowcolor{defaultcolor}
   \textbf{MemSVD} (MViTv2-S,  $16 \times 4$)  &    &  \textbf{43.0} & 64.5x3\x10 &  34.4\\        
	\end{tabular}
	\vspace{-.9em}
	\label{tab:sota:charades}
\end{table}

\section{Conclusion}
In this paper, we proposed an efficient mechanism for the modelling of a memory bank in long-term video understanding, which does not rely on the expensive attention mechanism. Instead, we proposed a low-rank SVD decomposition of the stored memory features and showed that an equivalent to attention can be obtained with a projection and reconstruction into the low-rank basis. We observe that similar or superior performance can be achieved with similar or even lower complexity than memory-based approaches. We note that such replacement brings a) complexity reduction by one order of magnitude, b) it matches or surpasses the accuracy achieved by the attention-based mechanisms and c) it can be seamlessly integrated with different architectures and input configurations. We will release the code and models for reproducibility. 

\vfill\pagebreak

\bibliographystyle{IEEEbib}
\bibliography{strings,refs}

\end{document}